\documentclass{article}
\pdfoutput=1

\usepackage[preprint,nonatbib]{neurips_2025}

\usepackage[T1]{fontenc}
\usepackage{amsmath}
\usepackage{microtype}
\usepackage{graphicx}
\usepackage{booktabs}
\usepackage{caption}
\usepackage{subcaption}
\usepackage[hidelinks]{hyperref}

\makeatletter
\renewcommand{\@noticestring}{Preprint.}
\makeatother

\newcommand{\num}[1]{#1}

\title{Grading Needs a Rubric, Not Intelligence}
\author{Jhen-Ke Lin\\
  National Yang Ming Chiao Tung University\\
  \texttt{jacob.cs14@nycu.edu.tw}}

\begin{document}

\maketitle

\begin{abstract}
\noindent
Small language models can grade open-ended examination answers as reliably as
substantially more expensive models when they grade against an explicit
rubric. We test this claim as the design principle behind
\textsc{any-to-bench}: a frontier model reads source documents once, at
ingestion, to extract each question and its rubric; lower-cost models then
perform all repeated grading work. We evaluate six
cost-efficient model configurations from two model families at three
reasoning-effort levels. Each configuration answers 24 open-ended examination
questions, and each also grades every answer sheet three times, yielding
3{,}456 per-question grades. Scores depend overwhelmingly on the answer
being graded: answer identity explains 95.6\,\% of score variance, whereas
judge identity explains only 0.2\,\%. Raising a writer's reasoning effort
moves earned scores by as much as 0.143 of full marks, while raising a
judge's reasoning effort moves assigned scores by at most 0.006. Six
frontier-tier judges, added as a check, reproduce these scores and are no
more reliable as a panel. Two
ablations then decompose the rubric on the same questions and answers. Removing its criteria and levels while keeping the official
answer changes nothing measurable. Removing the official answer as well
collapses reliability (ICC 0.888 to 0.628), inflates scores, and makes judge
reasoning effort matter again. The rubric is what decouples grading from
judge intelligence, and within the rubric the official answer does nearly
all the work. We find no evidence of length preference or same-family
preference under rubric-anchored grading.
\end{abstract}

\section{Introduction}

Turning an existing examination into a machine-runnable benchmark is mostly a
solved problem for multiple-choice items. It is an expensive problem for
everything else. Open-ended questions, such as proofs, essays, translations,
and drawings, need a grader, and the grader has historically been the costly,
unscalable part. Using a language model as the judge~\cite{zheng2023,chiang2023}
moved that cost from humans to models. But the standard recipe still spends a
frontier-model call on every grade, and it carries documented judge biases: a
preference for longer answers~\cite{dubois2024}, for answers in certain
positions~\cite{wang2023}, and for the judge's own model
family~\cite{panickssery2024}.

\textsc{any-to-bench} takes a different approach, built on an asymmetry
between its two model-driven steps: ingestion and grading. Ingestion is where
the intelligence goes. A frontier configuration reads the source documents and
extracts questions, answer formats, and scoring rubrics, and it does this
once per exam. Throughout this paper, a question's \emph{rubric} is its
complete extracted scoring standard: the official reference answer, plus
explicit criteria with defined levels where the source publishes them. Grading happens on every benchmark run, and the tool
bets that it needs almost no intelligence at all: with the rubric in hand,
grading reduces to applying it. That is work small models should handle
cheaply, repeatedly, and interchangeably.

That is a testable claim, and we test it. Six configurations served
both as answer writers and as judges over 24 open-ended questions from three
Taiwanese national examinations. They comprise the small tier of one model
family and the mid tier of another, each at three reasoning-effort settings.
Because the same configurations act both as writers and as judges, the
experiment turns the reasoning-effort dial once on each side of the grading
relation. The writer side shows what happens when capability matters; the
judge side tests whether it matters for grading. We find:

\begin{itemize}\setlength\itemsep{0.1em}
\item \textbf{Judge intelligence does not matter} (\S\ref{sec:asym}).
Reasoning effort moves what a writer \emph{earns} by up to 0.143 of full
marks, so capability registers when it sits on the answering side. The same
dial moves what a judge \emph{gives} by at most 0.006. Judge identity
accounts for 0.2\,\% of score variance, and replacing the entire panel with
its cheapest members changes per-answer scores by 0.019 on average.
Frontier-tier judges reproduce the same scores.
\item \textbf{The rubric is the mechanism, and its answer key is the
load-bearing part} (\S\ref{sec:rubric}). Two ablations decompose the rubric
on the same questions and answers. Stripping its criteria and levels while
keeping the official answer changes nothing measurable. Stripping the
official answer as well collapses reliability, inflates scores, and makes
judge effort matter again.
\item \textbf{Two textbook biases fail to appear} (\S\ref{sec:bias}). Under
rubric anchoring there is no self-preference and no length premium. A
within-question length--score correlation of $+0.6$ to $+0.7$ suggests
otherwise; the cross-writer comparison shows why that reading is wrong.
\item \textbf{One judge is enough} (\S\ref{sec:panel}). Mean panel reliability
is flat from two judges to six, and repeated grading adds nothing measurable.
What a larger panel adds is insurance against an unlucky pick of judge.
\end{itemize}

Throughout, the scale is anchored by construction rather than assumption
(\S\ref{sec:validity}). An empty answer sheet and the published reference
answers bound the scale from below and above, and every claim about agreement
is conditioned on the judges first getting both of these anchors right.

\section{Related work}

Reliability of LLM judges has mostly been studied in the pairwise-preference
setting, where a judge compares two answers. MT-Bench and Chatbot Arena report
over 80\,\% agreement between GPT-4 verdicts and human
preferences~\cite{zheng2023}. G-Eval aligns form-filling GPT-4 evaluation with
human judgments of generated text~\cite{liu2023}, and Chiang and
Lee~\cite{chiang2023} examine LLMs as replacements for human evaluation of
open text. The same literature documents the failure modes: pairwise judges
are sensitive to answer order~\cite{wang2023}, reward length until it is
explicitly controlled for~\cite{dubois2024}, and favour their own
generations~\cite{panickssery2024}. Surveys and systematic bias audits now
catalogue these and further failure modes at scale~\cite{gu2024,ye2025}.

Our setting differs in a way that matters for all three biases. The judge does
not compare two answers or rate free-form quality. It assigns points against
an extracted rubric with defined criteria and levels, on questions that carry
official point values. This is the classical instrument of educational
measurement, and we evaluate it with the classical tools: intraclass
correlation for absolute agreement~\cite{shroutfleiss1979} and a variance
decomposition over answers, judges, and their interaction. The length analysis
in \S\ref{sec:bias} is an instance of aggregation reversal~\cite{simpson1951},
also known as Simpson's paradox.

Rubric-based LLM grading of examinations is also studied directly. Recent
work grades nationwide school-leaving essays against curriculum rubrics and
reports agreement comparable to human panels~\cite{karjus2026}, and asks how
much rubric detail automated scoring needs~\cite{yoshida2025}. That line of
work varies the rubric and compares against human raters. We hold the rubric
fixed and vary the judge, asking how much intelligence rubric application
needs.

\section{Experimental design}
\label{sec:design}

\subsection{Corpus and question selection}

The corpus is 164 exam bundles (7{,}121 questions) that \textsc{any-to-bench}
produced from the published files of three Taiwanese national examinations:
the General Scholastic Ability Test (GSAT), the Advanced Subjects Test (AST),
and the unified entrance examination for technological and vocational
education (TVE), for the years 113--115 (2024--2026). The corpus is
public.\footnote{\url{https://huggingface.co/datasets/JacobLinCool/taiwan-exams}}
Ingestion is the one step that uses a frontier model: GPT-5.6 Sol at
extra-high (\texttt{xhigh}) reasoning effort read the source documents and
extracted each question, its answer format, and a per-criterion grading
rubric.

From the judge-graded questions in this corpus we drew 24, grouped into four
strata by how much scoring guidance the source itself publishes
(Table~\ref{tab:strata}, left half). Each stratum holds six questions. The 24
questions come from 21 different exam papers across the three series and the
three years, and cover eight subjects: Chinese, English, mathematics,
physics, chemistry, biology, geography, and civics. They are worth 1 to 25
points each and use three answer formats: ten short answers (including
English translation items), ten essays, and four drawings. Rubric size runs
from a single two-level criterion to rubrics with up to four criteria and up
to 26 levels in total. Appendix~\ref{app:questions} lists every question. The strata are not four arms of one experiment.
Stratum~B's two-level rubrics make agreement largely automatic, so it serves
as a floor check. Stratum~A, with no published rubric, is the comparison
condition that \S\ref{sec:rubric} turns on.

\subsection{Writers, anchor sheets, judges}

Six configurations wrote answers: GPT-5.6 Luna, the small tier of its model
family, and Claude Sonnet 5, the mid tier of its family, each at low, medium,
and high reasoning effort. Below we shorten these names to 5.6~Luna and
Sonnet~5.

Two further answer sheets were constructed rather than generated. We call them
\emph{anchors} because they pin the ends of the scale: a \emph{reference}
sheet containing each question's published official answer, which should earn
nearly full marks, and an \emph{empty} sheet, which should earn nothing.
Without human raters, these anchors are the only available ground truth. Eight
of the 24 questions publish no official answer (their examining boards do not
release keys for essays and translations). For those eight, the full-credit
anchor is undefined, so we exclude them from reference-anchor statistics only.

The same six configurations then served as judges. Each judge graded each of
the eight sheets three times: $8 \times 6 \times 3 = 144$ grading passes. One
graded question is one \emph{verdict}; the experiment produced 3{,}456
verdicts, all complete. To grade, the judge reads the question (with its
figures supplied as images), the rubric, and the answer, and assigns each
rubric criterion one of its defined levels. Drawing questions are answered as
precise textual descriptions of the drawing (shapes, labels, positions), and
the judge grades the description against the rubric. We analyse scores as
$p = \text{awarded}/\text{maximum}$, so a 25-point essay and a 1-point short
answer weigh alike. A difference of 0.10 in $p$ corresponds to 10 points on a
100-point scale.

Six further configurations at the frontier tier of each family also served
as judges: GPT-5.6 Sol, which performed ingestion, and Claude Opus~5, at the
same three efforts. Each graded all eight sheets once, as a direct check on
the pool's ceiling (\S\ref{sec:asym}). With the ablations of
\S\ref{sec:ablations}, the experiment totals 5{,}760 verdicts.

One design detail protects the results from our own tooling.
\textsc{any-to-bench} snaps each judged criterion score onto the rubric's
defined levels. Snapping mechanically increases apparent agreement: judges who
say 1.7 and 2.3 both become 2.0. We therefore recorded every pre-snap verdict
and computed all statistics both ways. Snapping altered \num{1 of 3{,}456}
verdicts, so no result below is an artifact of our own rounding.

\subsection{Measures}

We measure agreement with the intraclass correlation coefficient, form
ICC(2,1): two-way random effects, absolute agreement, single
measurement~\cite{shroutfleiss1979}. In plain terms, ICC(2,1) is the share of
score variance that reflects real differences between answers rather than
differences between judges. A value of 1 means the judges are interchangeable;
a value of 0 means the score depends on who graded rather than on what was
written. With
$n$ answers, $k$ judges, and the usual mean squares,
\begin{equation}
\text{ICC}(2,1) \;=\;
\frac{MS_R - MS_E}
     {MS_R + (k-1)\,MS_E + \tfrac{k}{n}\,(MS_C - MS_E)},
\label{eq:icc}
\end{equation}
where $MS_R$, $MS_C$, $MS_E$ are the answer, judge, and residual mean squares.
We use absolute agreement rather than consistency because a benchmark needs
judges to assign the \emph{same score}; agreement on ranking alone would be
too weak.
Confidence intervals come from a cluster bootstrap that resamples whole
questions, because verdicts on the same question are not independent. The same
mean squares give the variance decomposition
$\hat\sigma^2_{\text{answer}} = (MS_R - MS_E)/k$,
$\hat\sigma^2_{\text{judge}} = (MS_C - MS_E)/n$,
$\hat\sigma^2_{\text{res}} = MS_E$.

Unless stated otherwise, agreement statistics use a single grading pass,
because nobody grades three times in production. They also cover the six
model-written sheets only: the two anchors are trivially easy to separate and
would inflate any statistic that included them.

\subsection{Two ablations}
\label{sec:ablations}

The strata compare different questions, so they cannot separate the rubric's
effect from the questions' own character. Two ablations remove that limit by
intervening on the same questions. From the twelve stratum-C and -D questions
we built two further bundles. The first strips each rubric's criteria and
levels but keeps the official answer, which is the shape stratum-A rules
have. The second strips the official answer as well, leaving the judge
nothing but the question and the answer to grade. The same six judges then
re-graded all eight answer sheets once under each condition: 96 further
grading passes and 1{,}152 further verdicts. Comparisons against the full
rubric use the first repeat of the main study, so all three conditions are
single-pass.

\section{The scale is anchored}
\label{sec:validity}

\begin{figure}[!ht]
\centering
\includegraphics[width=0.9\linewidth]{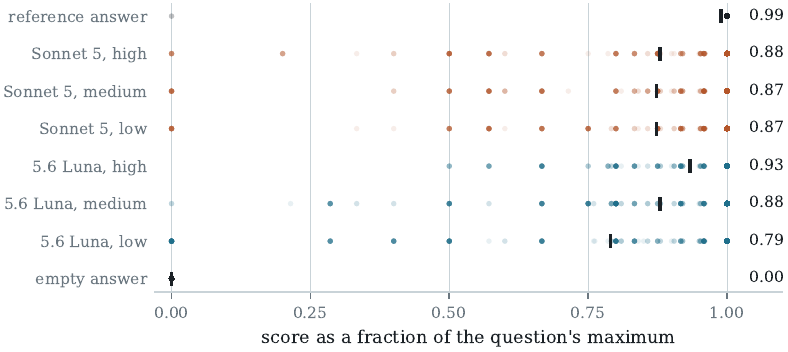}
\caption{All 3{,}456 verdicts, by answer sheet. The constructed anchors sit at
the ends of the scale, where they belong; the six writers sit strictly between
them. 5.6~Luna's answers improve with reasoning effort; Sonnet~5's do not, on
these questions.}
\label{fig:validity}
\end{figure}

The judges got both anchors right, almost without exception. This matters
because agreement alone proves nothing: judges can agree and all be wrong. The
empty sheet scored exactly zero in every one of its \num{432} gradings. The
reference sheet averaged \num{0.990} of full marks and scored at least 0.9 in
\num{285 of 288} gradings.\footnote{All three misses were single-pass flukes:
each was a lone zero, and in every case the same judge gave the same answer
full marks on both other repeats.} Figure~\ref{fig:validity} shows the two anchors
bracketing all six writers, with every verdict drawn.

\section{Effort moves writers, not judges}
\label{sec:asym}

\begin{figure}[!ht]
\centering
\includegraphics[width=0.9\linewidth]{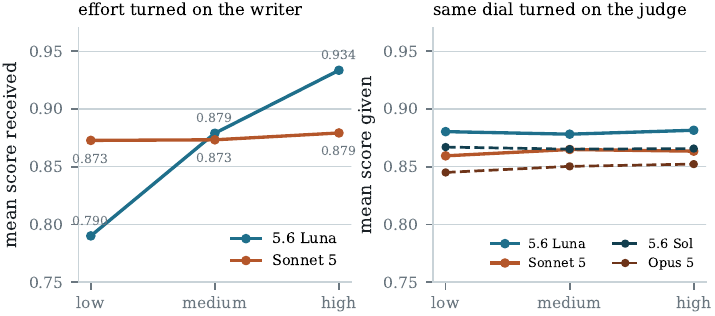}
\caption{The same dial, both sides. Left: mean score \emph{received} by each
family's answers as writer effort rises. Right: mean score \emph{given} by
each family's judges as judge effort rises, over the identical answers;
dashed lines are the frontier tier, graded once. Both panels span the same
0.22 of the scale; on equal axes, the judge side stays flat.}
\label{fig:dial}
\end{figure}

Reasoning effort changes what an answer earns. It does not change what a judge
gives. Because the six configurations act as both writers and judges, both
effects come from the same experiment, and Figure~\ref{fig:dial} shows them on
a common scale. Turned on the writer, the effort dial moves earned scores by
\num{0.143} of full marks for 5.6~Luna (0.790 to 0.934). Sonnet~5's answers
are flat in effort, so the writer-side response is carried by one family; one
is enough to show that the instrument registers capability whenever
capability differs. Turned on the judge, the same dial moves the mean verdict
by \num{+0.001} (5.6~Luna) and \num{+0.004} (Sonnet~5). The six judges'
mean leniencies, across two families and three efforts, all sit within
$\pm 0.012$ of the panel mean.

Swapping the whole panel makes the judge-side null concrete. We replaced the
two most expensive judges (high effort, both families) with the two cheapest
(low effort). Per-answer scores changed by \num{0.019} on average (maximum
0.167, rank correlation 0.90). The cheap pair's single-pass reliability,
ICC(2,1) $=$ \num{0.913}, is statistically indistinguishable from the full
six-judge panel's \num{0.922} (95\,\% confidence interval 0.815--0.966,
cluster bootstrap over questions). In the variance decomposition over all
writer answers, which answer is being graded explains \num{95.6\,\%} of score
variance. Which judge is grading explains \num{0.2\,\%}. The remaining
4.3\,\% is answer--judge interaction.\footnote{Computed on three-pass mean
scores, so the residual reflects interaction rather than repeat noise.}

Effort does improve one thing on the judge side: repeatability. When re-grading
the identical answer, the low-effort 5.6~Luna judge varies with a
within-judge standard deviation of 0.080, against 0.038 at high effort
(Sonnet~5: 0.038 against 0.026). But this variation is centred on the same verdict, and it is
already included in the single-pass figures above. It changes no comparison
between answers. Nowhere in 3{,}456 verdicts does effort change what an
answer is worth.

The main judge pool is deliberately downmarket: the small tier of one
family and the mid tier of another. As a direct check on its ceiling, six
frontier-tier configurations, GPT-5.6 Sol (the ingestion model's own tier)
and Claude Opus~5 at the same three efforts, graded every sheet once. They
change nothing (Table~\ref{tab:judges}). Each frontier judge deviates from
the cheap panel's consensus by 0.030 to 0.039, strictly inside the cheap
judges' own leave-one-out range (0.023 to 0.042). Their effort dial is as
flat as the cheap judges'
(Figure~\ref{fig:dial}, dashed lines). Their anchors hold, with one
reference miss in 96 gradings. And the most expensive pair available, 5.6~Sol
and Opus~5 at high effort, is slightly \emph{less} reliable as a panel (ICC
0.813) than the two cheapest judges (0.913). From 5.6~Luna at low effort to
Opus~5 at high, judge capability buys grading nothing. This confirms the
grading half of the tool's asymmetry directly: the intelligence was spent at
ingestion, and a more intelligent judge has nothing left to improve.

\begin{table}[!ht]
\centering\small
\caption{All twelve judges, single pass, over the six writer sheets.
\emph{Deviation} is the mean absolute difference from the cheap panel's
consensus on the same answer; for the six cheap judges the consensus
excludes their own verdict. The frontier tier sits inside the cheap tier's
own range on both columns.}
\label{tab:judges}
\begin{tabular}{@{}llcc@{}}
\toprule
\textbf{Tier} & \textbf{Judge} & \textbf{Mean score given} & \textbf{Deviation} \\
\midrule
cheap    & 5.6 Luna low     & 0.873 & 0.032 \\
cheap    & 5.6 Luna medium  & 0.889 & 0.042 \\
cheap    & 5.6 Luna high    & 0.881 & 0.035 \\
cheap    & Sonnet 5 low     & 0.864 & 0.030 \\
cheap    & Sonnet 5 medium  & 0.864 & 0.030 \\
cheap    & Sonnet 5 high    & 0.862 & 0.023 \\
\midrule
frontier & 5.6 Sol low      & 0.867 & 0.030 \\
frontier & 5.6 Sol medium   & 0.865 & 0.036 \\
frontier & 5.6 Sol high     & 0.866 & 0.039 \\
frontier & Opus 5 low       & 0.845 & 0.036 \\
frontier & Opus 5 medium    & 0.850 & 0.034 \\
frontier & Opus 5 high      & 0.852 & 0.032 \\
\bottomrule
\end{tabular}
\end{table}

\section{The rubric carries the judgment}
\label{sec:rubric}

Why can small judges grade? Stratum~A shows what happens when the rubric is
absent. Its six questions come from sources that publish no rubric, so
ingestion could extract only a reference answer or less, and the judge must
supply the judgment itself. Table~\ref{tab:strata} shows the result.

\begin{table}[!ht]
\centering\small
\caption{Strata design (left) and outcomes (right). \emph{Judge disagreement}
is the mean spread (max$-$min) among the six judges on the same answer.
\emph{Answer spread} is the standard deviation of per-answer mean scores:
how far apart the judges are able to place the six writers' answers.
Stratum~B's agreement is largely guaranteed by its two-level rubrics;
stratum~A is the no-rubric comparison condition.}
\label{tab:strata}
\begin{tabular}{@{}clc@{\hspace{1.5em}}ccc@{}}
\toprule
& \textbf{Scoring guidance in source} & \textbf{In corpus}
& \textbf{ICC(2,1)} & \textbf{Judge disagr.} & \textbf{Answer spread} \\
\midrule
A & none, or reference answer only        & 32  & 0.466 & 0.080 & \num{0.043} \\
B & one criterion, two levels             & 105 & 0.983 & 0.019 & 0.276 \\
C & one criterion, three or more levels   & 116 & 0.953 & 0.092 & 0.221 \\
D & two or more criteria                  & 57  & 0.944 & 0.085 & 0.233 \\
\bottomrule
\end{tabular}
\end{table}

\begin{figure}[!ht]
\centering
\includegraphics[width=0.9\linewidth]{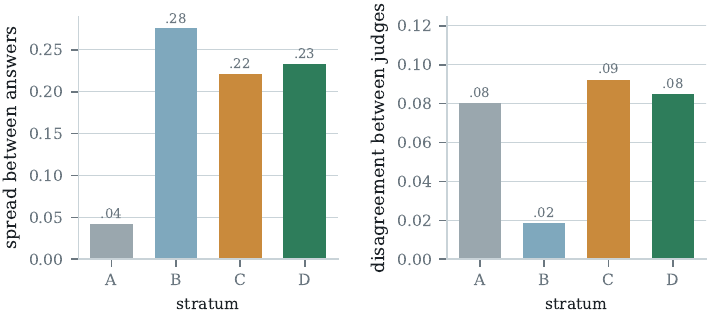}
\caption{The stratum contrast. Left: how far apart the six writers' answers
are placed, by stratum. Right: how much the six judges disagree on the same
answer. Stratum~A's judges are no more divided than C's or D's; they simply
have nothing to divide.}
\label{fig:discriminate}
\end{figure}

Read naively, the table says judges agree \emph{worse} without a rubric: A's
ICC is 0.466 against 0.94--0.98 elsewhere. The naive reading is wrong, and the
error is worth explaining because reliability coefficients are how this
literature reports results. ICC is signal over signal-plus-noise. Stratum~A's
judges disagree with each other about as much as C's and D's in absolute terms
(0.080 against 0.092 and 0.085); the noise is ordinary. What has collapsed is
the signal. In stratum~A the judges place the six answers within a standard
deviation of \num{0.043} of each other, five times narrower than the
0.22--0.28 seen elsewhere (Figure~\ref{fig:discriminate}). Every answer gets
much the same mark.\footnote{The same collapse explains why stratum~A's
within-judge repeat variation (0.019) is the \emph{lowest} of the four
strata: a judge that gives everything 0.85 is very stable.}

The stratum contrast, however, compares different questions, and it is
confounded: stratum~A's questions are mostly free-form writing, and prose
may compress for reasons of its own. The two ablations of
\S\ref{sec:ablations} settle what the rubric itself contributes, because
they intervene on the same questions and the same answers.

\begin{figure}[!ht]
\centering
\includegraphics[width=0.82\linewidth]{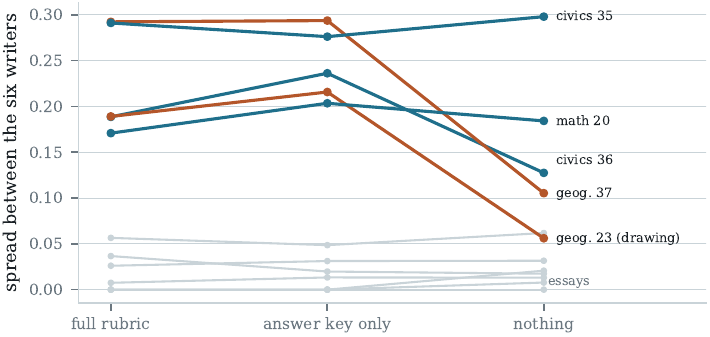}
\caption{Writer spread per question across the guidance spectrum. Removing
the criteria and levels (middle) changes nothing. Removing the official
answer as well (right) collapses the questions that can only be checked
against it, while the re-derivable mathematics and civics questions survive.
The essays sit flat at every guidance level.}
\label{fig:ablate}
\end{figure}

Stripping the criteria and levels changes nothing measurable
(Figure~\ref{fig:ablate}, middle). On the five questions with real
discrimination, the spread between writers keeps a median 114\,\% of its
full-rubric value. Reliability is unchanged: ICC 0.880 with the full rubric,
0.888 with the official answer alone. The judges become slightly more
generous (+0.016 of full marks) and slightly less aligned on each answer
(mean spread 0.122 to 0.145), and that is all. Given the official answer,
the elaborate criteria are redundant.

Stripping the official answer as well breaks things
(Figure~\ref{fig:ablate}, right). Reliability falls to 0.628. Scores inflate
by 0.074 of full marks. The reference anchor drops below perfection for the
first time in this paper (0.957). Median discrimination falls to 68\,\%, and
which questions survive is telling. The two questions a judge can re-derive
for itself hold their spread: the mathematics question keeps 108\,\% and one
civics question 102\,\%. The questions whose answers can only be checked against
the key collapse to 30--36\,\%. Judge reasoning effort, irrelevant
everywhere else in this paper, also starts to matter: without the key, every
judge drifts further from the full-rubric consensus, and the drift now falls
with effort, 0.143 at low 5.6~Luna effort against 0.114 at high, twice the
gradient seen when the key is present.

This locates the mechanism. The rubric is what decouples grading from judge
intelligence: without it, grading turns back into answering, and capability
matters again. Within the rubric, the official answer carries nearly all of
the weight, and the criteria and levels add a margin of consistency. This is
the ingestion half of the asymmetry made concrete: the frontier model's real
product is the scoring standard, and above all the key.

One limit holds at every guidance level. The essays never discriminate: with
a 26-level rubric, with only a key, or with nothing, the six writers' essays
sit within 0.04 of one another. Whether the essays are genuinely that
similar or the judges cannot separate prose quality is not decidable without
human raters (\S\ref{sec:limits}).

\section{Two biases that fail to appear}
\label{sec:bias}

\subsection{Length, and the trap in measuring it}

Within a single question, longer answers score higher. The median
within-question rank correlation between answer length and score is $+0.60$ to
$+0.74$ for every one of the six judges. Reported alone, that number reads as
a length premium~\cite{dubois2024}. Reported alone, it would mislead.

The cross-writer comparison contradicts it (Figure~\ref{fig:twobias}, left).
The six answer sheets fall into two length regimes: the three 5.6~Luna
sheets stay near 230 characters, while the three Sonnet~5 sheets run
2.3$\times$ longer, near 520. If judges paid for characters, the long regime would
outscore the short one. It does not: the family means are 0.875 and 0.868,
essentially equal. Meanwhile, the experiment's entire quality range, 0.790 to
0.934, occurs \emph{within} the short regime, at nearly constant length. At
matched high effort, the 234-character writer outscores the 538-character
writer by 0.055. Length varies by a factor of 2.3 with no effect on scores;
quality varies at fixed length with full effect.

The within-question correlation is therefore an instance of aggregation
reversal~\cite{simpson1951}. Within a question, the answer that covers more of
the rubric's criteria is both longer and better, so length stands in for
completeness. Across writing styles, the stand-in breaks. What a rubric-anchored judge
pays for is the criteria an answer covers.

\subsection{Self-preference}

\begin{figure}[!ht]
\begin{subfigure}[t]{0.46\linewidth}
\centering
\includegraphics[width=\linewidth]{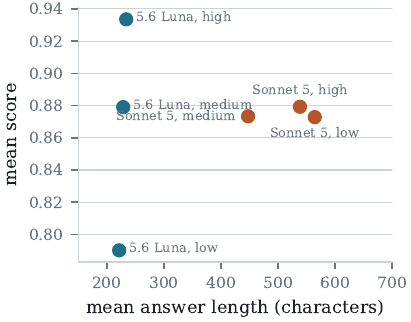}
\caption{Answer length against mean score, per sheet. The long regime gains
nothing; the short regime spans the whole quality range.}
\label{fig:length}
\end{subfigure}\hfill
\begin{subfigure}[t]{0.5\linewidth}
\centering
\includegraphics[width=\linewidth]{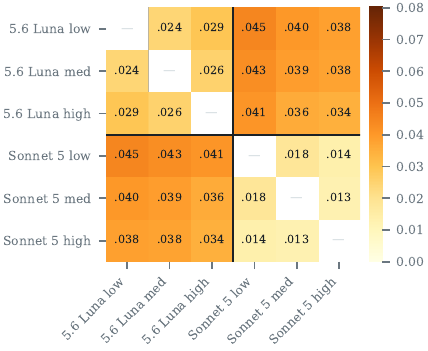}
\caption{Mean disagreement for every judge pair; rules mark the family
boundary.}
\label{fig:pairs}
\end{subfigure}
\caption{Two biases the pairwise-preference literature warns about, absent
under rubric anchoring.}
\label{fig:twobias}
\end{figure}

5.6~Luna judges score 5.6~Luna answers at 0.874 and Sonnet~5 answers at
0.886. Sonnet~5 judges score them at 0.861 and 0.864. Each family is, if anything, marginally
kinder to the \emph{other} family's prose, and the two gaps run in opposite
directions, which is what noise looks like~\cite{panickssery2024}. The
family boundary is visible but faint in the pairwise disagreement structure
(Figure~\ref{fig:twobias}, right): judges disagree by 0.021 within a family
and 0.039 across, small in both cases. One caveat applies. With one model per
family, self-preference cannot be separated from model identity, so this
result shows the bias absent under rubric anchoring here; it does not rule
the bias out in general.

\section{One judge is enough}
\label{sec:panel}

\begin{figure}[!ht]
\centering
\includegraphics[width=0.5\linewidth]{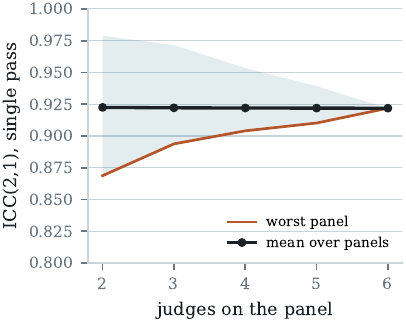}
\caption{Reliability against panel size, over all $\binom{6}{k}$ panels. The
mean is flat from $k=2$; what grows with panel size is the guaranteed
minimum.}
\label{fig:panel}
\end{figure}

Multi-judge panels are standard practice, so we measured their effect here:
almost nothing. Averaged over every possible panel, single-pass reliability is
\num{0.923} with two judges and \num{0.922} with six
(Figure~\ref{fig:panel}). Even the \emph{worst} two-judge panel reaches 0.869,
and adding judges narrows the worst case rather than raising the mean.
Repetition helps as little: 73\,\% of repeated gradings are exactly identical,
the pooled within-judge standard deviation is 0.050, and averaging three
passes leaves every headline number unchanged. A second judge is inexpensive
insurance against an unlucky first choice. A sixth only adds cost.

\section{Limitations}
\label{sec:limits}

\textbf{The floor.} The frontier comparison closes the question upward:
5.6~Sol and Opus~5 grade no better than 5.6~Luna and Sonnet~5. Downward it stays
open: the cheap judges are small only relative to their frontier siblings,
and nothing here shows how far down the capability scale rubric application
remains intact.

\textbf{Scope.} All 24 questions come from Taiwanese national examinations
and are graded in Traditional Chinese; other languages and examining
traditions are untested. The best writer averages 0.934, close enough to the
ceiling that discrimination between two \emph{good} answers is
under-represented. No human marker scored these answers: \emph{valid} here
means anchored and consistent, and says nothing about agreement with a human
examiner.

\textbf{Power.} The headline reliability is adequately powered; the bootstrap
lower bound, 0.815, still supports strong agreement, and no single question
moves any stratum's ICC by more than 0.17. But six questions per stratum
remains thin. The strata are observational rather than randomised; the
ablations remove this limit for the rubric's own effect, but they are
single-pass and cover twelve questions. And the writer-side effort response
rests on one family, since the other family's answers were flat in effort.

\textbf{Prose.} The essays never discriminate, at any guidance level. Six
competent models may genuinely write prose of similar merit, or the judges
may be unable to separate prose quality; deciding between the two requires
human raters. Either way, benchmark builders should expect little signal
from free-form writing items.

\section{Conclusion}

An examination stores judgment that someone has already exercised: what to
ask, what a good answer contains, and how many points each part is worth.
\textsc{any-to-bench} recovers that judgment as a rubric, using a frontier
model once per exam. The experiment reported here says the recovery works and
the economics follow. After ingestion, grading is the easy half. Small judges
at low effort grade the same answers to the same scores as their most
expensive counterparts, contribute 0.2\,\% of score variance, show neither
the length bias nor the family bias reported for pairwise judging, and need
no panel. The one thing they cannot lose is the rubric. Without it, scores
inflate, judges drift apart, and reasoning effort begins to matter again:
grading turns back into answering. Within the rubric, the official answer
carries nearly all the weight. Intelligence belongs at ingestion. A
benchmark that spends it there is cheap to run at any scale.

\appendix

\section{The 24 questions}
\label{app:questions}

Table~\ref{tab:questions} lists every question in the study: its stratum, the
exam paper it comes from, its answer format, its point value, and the size of
its rubric.

\begin{table}[!ht]
\centering\footnotesize
\caption{The 24 questions. \emph{Levels} is the total number of defined
levels across the rubric's criteria; stratum~A questions publish no rubric.
\emph{Key} marks whether the examining board publishes an official answer;
the eight questions without one are excluded from reference-anchor
statistics.}
\label{tab:questions}
\setlength{\tabcolsep}{3.5pt}
\begin{tabular}{@{}clllrrrc@{}}
\toprule
& \textbf{Source exam} & \textbf{Question} & \textbf{Format}
& \textbf{Points} & \textbf{Criteria} & \textbf{Levels} & \textbf{Key} \\
\midrule
A & \texttt{ast-114-physics} & q19 & short answer & 4 & 0 & -- & yes \\
A & \texttt{tve-113-chinese} & q39 & essay & 24 & 0 & -- & no \\
A & \texttt{tve-113-english} & q44 & short answer & 6 & 0 & -- & no \\
A & \texttt{tve-113-foreign-language-english-2} & q8 & essay & 24 & 0 & -- & no \\
A & \texttt{tve-114-chinese} & q39 & essay & 24 & 0 & -- & no \\
A & \texttt{tve-114-foreign-language-english-2} & q9 & essay & 24 & 0 & -- & no \\
\midrule
B & \texttt{ast-113-biology} & q37 & short answer & 2 & 1 & 2 & yes \\
B & \texttt{ast-113-math-jia} & q17 & essay & 6 & 1 & 2 & yes \\
B & \texttt{ast-113-physics} & q25.a & short answer & 1 & 1 & 2 & yes \\
B & \texttt{ast-114-math-jia} & q12.b & essay & 4 & 1 & 2 & yes \\
B & \texttt{ast-114-math-yi} & q18 & drawing & 8 & 1 & 2 & yes \\
B & \texttt{ast-115-geography} & q20.b & drawing & 3 & 1 & 2 & yes \\
\midrule
C & \texttt{ast-113-civics} & q36.b & short answer & 5 & 1 & 6 & yes \\
C & \texttt{ast-113-geography} & q36 & drawing & 3 & 1 & 4 & yes \\
C & \texttt{ast-115-civics} & q35 & short answer & 6 & 1 & 7 & yes \\
C & \texttt{gsat-113-english} & q19.a & short answer & 4 & 1 & 9 & yes \\
C & \texttt{gsat-114-chinese-writing} & q2 & essay & 25 & 1 & 26 & no \\
C & \texttt{gsat-115-chinese-writing} & q1.b & essay & 21 & 1 & 22 & no \\
\midrule
D & \texttt{ast-113-chemistry} & q25 & short answer & 4 & 4 & 8 & yes \\
D & \texttt{ast-113-geography} & q37 & short answer & 5 & 2 & 6 & yes \\
D & \texttt{ast-114-geography} & q42.b & short answer & 6 & 2 & 6 & yes \\
D & \texttt{ast-115-geography} & q23.c & drawing & 7 & 2 & 6 & yes \\
D & \texttt{gsat-113-english} & q20 & essay & 20 & 4 & 24 & no \\
D & \texttt{gsat-115-math-a} & q20 & essay & 8 & 4 & 8 & yes \\
\bottomrule
\end{tabular}
\end{table}

\section{Reproducibility}
\label{app:repro}

Every number and figure in this paper regenerates from the stored grading
reports. All materials live in the \texttt{research/judge-reliability}
directory of the tool's
repository.\footnote{\url{https://github.com/JacobLinCool/any-to-bench}}
They are:

\begin{itemize}\setlength\itemsep{0.1em}
\item the 144 grading reports of the main study, one JSON file per (sheet,
judge, repeat);
\item the 96 ablation grading reports: \texttt{reports\_ab} for the
key-only condition and \texttt{reports\_bare} for zero guidance;
\item the 48 frontier-judge reports (\texttt{reports\_frontier});
\item \texttt{data.csv}, a tidy table with one row per main-study verdict
(3{,}456 rows);
\item the analysis and figure scripts, which read only the reports above.
\end{itemize}

The exam corpus, including the 24 selected questions and their rubrics, is
public at \url{https://huggingface.co/datasets/JacobLinCool/taiwan-exams}.
Table~\ref{tab:columns} defines the columns of \texttt{data.csv}.

\begin{table}[!ht]
\centering\small
\caption{Columns of \texttt{data.csv}. Each row is one verdict: one judge
grading one question on one answer sheet in one repeat.}
\label{tab:columns}
\begin{tabular}{@{}ll@{}}
\toprule
\textbf{Column} & \textbf{Meaning} \\
\midrule
\texttt{qid} & question identifier, prefixed by its source bundle \\
\texttt{stratum} & scoring-guidance stratum, A--D (Table~\ref{tab:strata}) \\
\texttt{max\_points} & the question's maximum points \\
\texttt{type} & answer format (short answer, essay, drawing) \\
\texttt{sheet} & answer sheet being graded: a writer sheet or an anchor \\
\texttt{judge} & judge configuration (family and reasoning effort) \\
\texttt{jfam}, \texttt{wfam} & judge family and writer family \\
\texttt{repeat} & grading repeat, 1--3 \\
\texttt{p} & awarded/maximum after level snapping \\
\texttt{p\_raw} & awarded/maximum before level snapping \\
\texttt{snapped} & 1 if snapping changed this verdict, else 0 \\
\texttt{len} & answer length in characters \\
\bottomrule
\end{tabular}
\end{table}

\end{document}